\documentclass[letterpaper]{article} % DO NOT CHANGE THIS
\usepackage{aaai2026}  % DO NOT CHANGE THIS
\usepackage{times}  % DO NOT CHANGE THIS
\usepackage{helvet}  % DO NOT CHANGE THIS
\usepackage{courier}  % DO NOT CHANGE THIS
\usepackage[hyphens]{url}  % DO NOT CHANGE THIS
\usepackage{graphicx} % DO NOT CHANGE THIS
\usepackage{natbib}  % DO NOT CHANGE THIS AND DO NOT ADD ANY OPTIONS TO IT
\usepackage{caption} % DO NOT CHANGE THIS AND DO NOT ADD ANY OPTIONS TO IT
\usepackage{algorithm}
\usepackage{algorithmic}
\usepackage{amssymb}
\usepackage{subcaption}
\usepackage{multirow}
\usepackage{amsmath}
\usepackage{amsthm}
\usepackage{booktabs}

\usepackage{newfloat}
\usepackage{listings}
\DeclareCaptionStyle{ruled}{labelfont=normalfont,labelsep=colon,strut=off} % DO NOT CHANGE THIS
\floatstyle{ruled}
\newfloat{listing}{tb}{lst}{}
\floatname{listing}{Listing}
\title{Learngene Search Across Multiple Datasets for Building Variable-Sized Models}
\author{
    Boyu Shi\textsuperscript{\rm 1,2},
    Junbo Zhou\textsuperscript{\rm 1,2},
    Chang Liu\textsuperscript{\rm 1,2},
    Xu Yang\textsuperscript{\rm 1,2,\ensuremath{*}},\\
    Qiufeng Wang\textsuperscript{\rm 1,2},
    Xin Geng\textsuperscript{\rm 1,2,\ensuremath{\dagger}}
}
\affiliations{
    \textsuperscript{\rm 1}School of Computer Science and Engineering, Southeast University, Nanjing, China\\
    \textsuperscript{\rm 2}Key Laboratory of New Generation Artificial Intelligence Technology and Its Interdisciplinary Applications (Southeast University), Ministry of Education, China\\
    \ensuremath{*}Co-corresponding author. \ensuremath{\dagger}Co-corresponding author.\\
    \{shiboyu,220242458,liuchang0520,xuyang\_palm,qfwang,xgeng\}@seu.edu.cn
}

\usepackage{bibentry}
\begin{document}

\maketitle

\begin{abstract}
    Deep learning methods are widely applied in various scenarios with diverse resource constraints, resulting in models of varying sizes, such as the Vision Transformer (ViT) series. Deploying these models typically requires pretraining from scratch followed by finetuning, which is time-consuming and computationally expensive. To address these challenges, the Learngene paradigm has been introduced. 
    This paradigm extracts key components, referred to as learngene, which possess strong learning capabilities, from a well-trained large model, known as the ancestry model (Ans-Net) and uses the learngene to initialize variable-sized models, called descendant models (Des-Nets).
    Previous methods for extracting learngene focused on a single dataset, which limited the effectiveness of the extraction and resulted in suboptimal performance, particularly in downstream tasks. To overcome this limitation, we propose a novel method for extracting learngene across multiple datasets, called Learngene Search Across Multiple Datasets for Building Variable-Sized Models (LSAMD).
    LSAMD expands the Ans-Net into a searchable super Ans-Net by introducing dataset-specific blocks and dataset adapters (DADs) in each layer. The dataset-specific blocks learn from the input dataset, while the original blocks in the Ans-Net, called the base blocks remain frozen to retain the initial knowledge. The DAD selects the propagation path between the dataset-specific and base blocks based on the input dataset. During training, LSAMD searches for an optimal architecture path for each dataset from the super Ans-Net. The base blocks most frequently adopted by these paths are then extracted as learngene for initializing variable-sized Des-Nets.
    Extensive experiments across multiple datasets demonstrate that LSAMD not only achieves performance comparable to the pretrain-finetune method but also significantly reduces both storage and training costs.
\end{abstract}

\section{Introduction}
Vision Transformers (ViTs) \cite{Dosovitskiy2020AnII} have made significant advancements in computer vision due to high performance. This capability enables their deployment across a wide range of scenarios, from resource-constrained mobile devices to resource-rich computing centers. The diversity of resource settings results in models of varying sizes. Typically, these models are initialized randomly and trained from scratch for deploying. However, the training cost associated with this approach increases substantially as the number of deployment scenarios grows.

To address this issue, the Pretrain-Finetune (PF) method has been proposed. This approach involves initially training a source model on a large-scale dataset, such as ImageNet-1K \cite{Deng2009ImageNetAL}, and subsequently finetuning the model on target datasets to derive the target model. However, this method requires transferring the entire source model to downstream tasks, which introduces several challenges.
For each resource constraint, PF trains a model of a specific size and fine-tunes it for downstream scenarios. As the number of resource constraints increases, the PF method becomes increasingly time-consuming. Besides, PF necessitates storing pretrained models of varying sizes for fine-tuning, leading to significant storage requirements and reduced flexibility for downstream tasks with diverse resource constraints.

\begin{figure*}[t]
    \centering
    \includegraphics[width=0.9\textwidth]{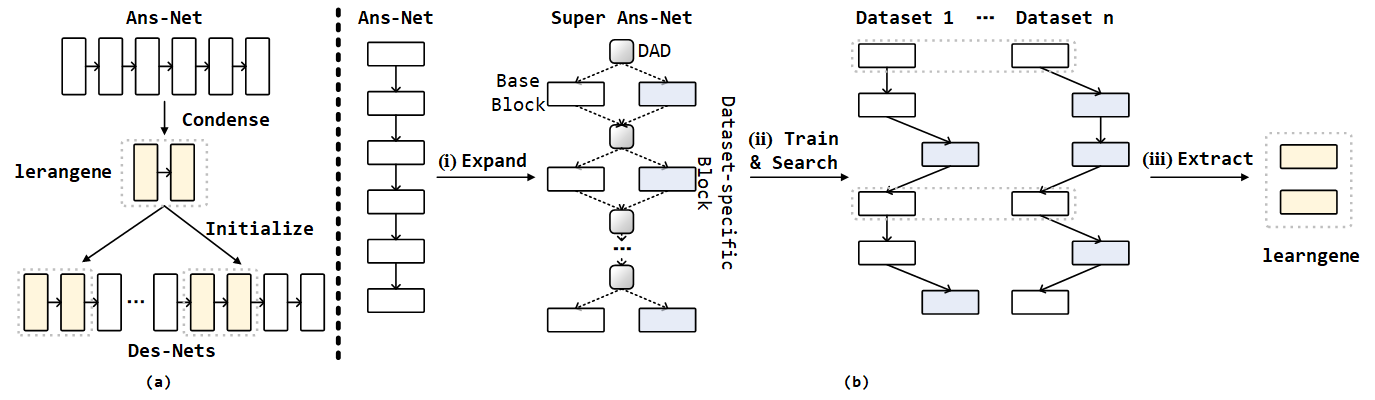}
    \caption[Figure]{(a) The framework of Learngene. The Ans-Net refers to the ancestry model, and the Des-Nets are descendant models. (b) The overall steps of extracting learngene in LSAMD. (\romannumeral1) We expand the Ans-Net into the super Ans-Net by adding extra dataset-specific blocks and dataset adapter (DAD). (\romannumeral2) The super Ans-Net is trained on multiple datasets and we search a specific path for each dataset. (\romannumeral3) Among these paths, the mostly used blocks from the Ans-Net are extracted as learngene. 
    }
    \label{fig:motivation}
\end{figure*}

Today, a novel learning paradigm called Learngene \cite{wang2022learngene,Wang2023LearngeneIC} has been introduced to handle these challenges. The paradigm has two main steps as shown in Figure \ref{fig:motivation}a: (1) Learngene first extracts a small but critical part called learngene from the well-trained large model termed as the ancestry model (Ans-Net). (2) The extracted learngene is then used to initialize variable-sized smaller models, termed descendant models (Des-Nets). Since containing the critical knowledge from the Ans-Net, the learngene has the capability to initialize Des-Nets to perform well on downstream datasets. Besides, the compact scale of learngene allows it to initialize a multitude of diverse medium-sized Des-Nets, thereby exhibiting considerable flexibility in adapting to downstream tasks with varying resource constraints. Since learngene is crucial for the development of Des-Nets, the extraction of appropriate learngene is important. To achieve this, several studies have been conducted to propose methods for extracting learngene from the Ans-Net. 

Specifically, Vanilla learngene (Van-LG) \cite{wang2022learngene} extracts learngene based on the gradient information of Ans-Net during training across various tasks derived from a single dataset, such as Cifar100 \cite{Krizhevsky2009LearningML} or ImageNet100 \cite{Deng2009ImageNetAL}. 
Additionally, Auto-Learngene (AL) \cite{Wang2023LearngeneIC} trains a meta-network to calculate the similarity between the output of each layer of the Ans-Net and the auxiliary models on Cifar100 \cite{Krizhevsky2009LearningML} or ImageNet100 \cite{Deng2009ImageNetAL}, and selects the most similar layer in the Ans-Net as learngene.
PEG \cite{wangvision} employs a probabilistic mixture approach to sample a subset of Multi-Head Self-Attention and Feed-Forward layers from Ans-Net as learngene, utilizing ImageNet-1K \cite{Deng2009ImageNetAL}.
% Cluster-Learngene extracts learngene by adaptively clustering critical internal modules, including attention heads and position-wise feed-forward networks (FFNs) of the Ans-Net.
LearngenePool \cite{shi2024building}, TLEG \cite{xia2024transformer}, SWS \cite{xia2024exploring} and LeTs \cite{xia2024initializing} distill the Ans-Net into one or more smaller auxiliary models using a single dataset, ImageNet-1K \cite{Deng2009ImageNetAL}, to extract learngene. 
However, these studies has the following limitation. They extract learngene from \textbf{a single dataset}. One dataset typically follows a specific data distribution, which limits the definition of critical part with general knowledge from the Ans-Net and results in reduced performance of Des-Nets initialized by learngene.

Motivated by this problem, we propose a new method called \textbf{L}earngene \textbf{S}earch \textbf{A}cross \textbf{M}ultiple \textbf{D}atasets (LSAMD) to extract learngene from the Ans-Net on multiple datasets. The proposed LSAMD involves the following stages. As shown in Figure \ref{fig:motivation}b (\romannumeral1), we first \textbf{expand} the architecture of the Ans-Net into a super Ans-Net. In each layer of the super Ans-Net, it additionally incorporates a dataset-specific block and a Dataset Adapter (DAD). The dataset-specific block is initialized with the same parameters as the original block of the Ans-Net, termed as the base block. The DAD determines the forward propagation path between base and dataset-specific blocks when training the super Ans-Net. 
Then, as shown in Figure \ref{fig:motivation}b (\romannumeral2), we train the super Ans-Net on multiple datasets. During this process, dataset-specific blocks are updated while base blocks remain fixed, which allows each layer to either learn dataset-specific knowledge or utilize the existing capabilities of Ans-Net. 
After training, DAD \textbf{searches} a specific propagation path for each dataset from the super Ans-Net. We then conduct a layer-wise statistical analysis of the usage frequency of base blocks across all searched paths. Based on this, each base block in the Ans-Net is categorized into one of two scenarios: heavily used, or lightly used. The heavily used base blocks are identified as the critical modules shared across datasets and \textbf{extracted} as learngene, as illustrated in Figure \ref{fig:motivation}b (\romannumeral3).
Finally, we utilize the extracted learngene to initialize variable-sized models for downstream tasks with various resource constraints.

Our main contributions are as follows: (1) We extend learngene extraction from a single dataset to multiple datasets. To the best of our knowledge, this is the first work in the Learngene paradigm to consider a multi-dataset scenario. (2) We introduce a method called LSAMD for searching learngene from the expanded Ans-Net across multiple datasets. (3) Extensive experiments demonstrate that LSAMD can flexibly build variable-sized models, achieve performance comparable to the pretrain-finetune approach while using less storage and reduce training costs on the downstream tasks.

\begin{figure*}[t]
    \centering
    \includegraphics[width=0.9\textwidth]{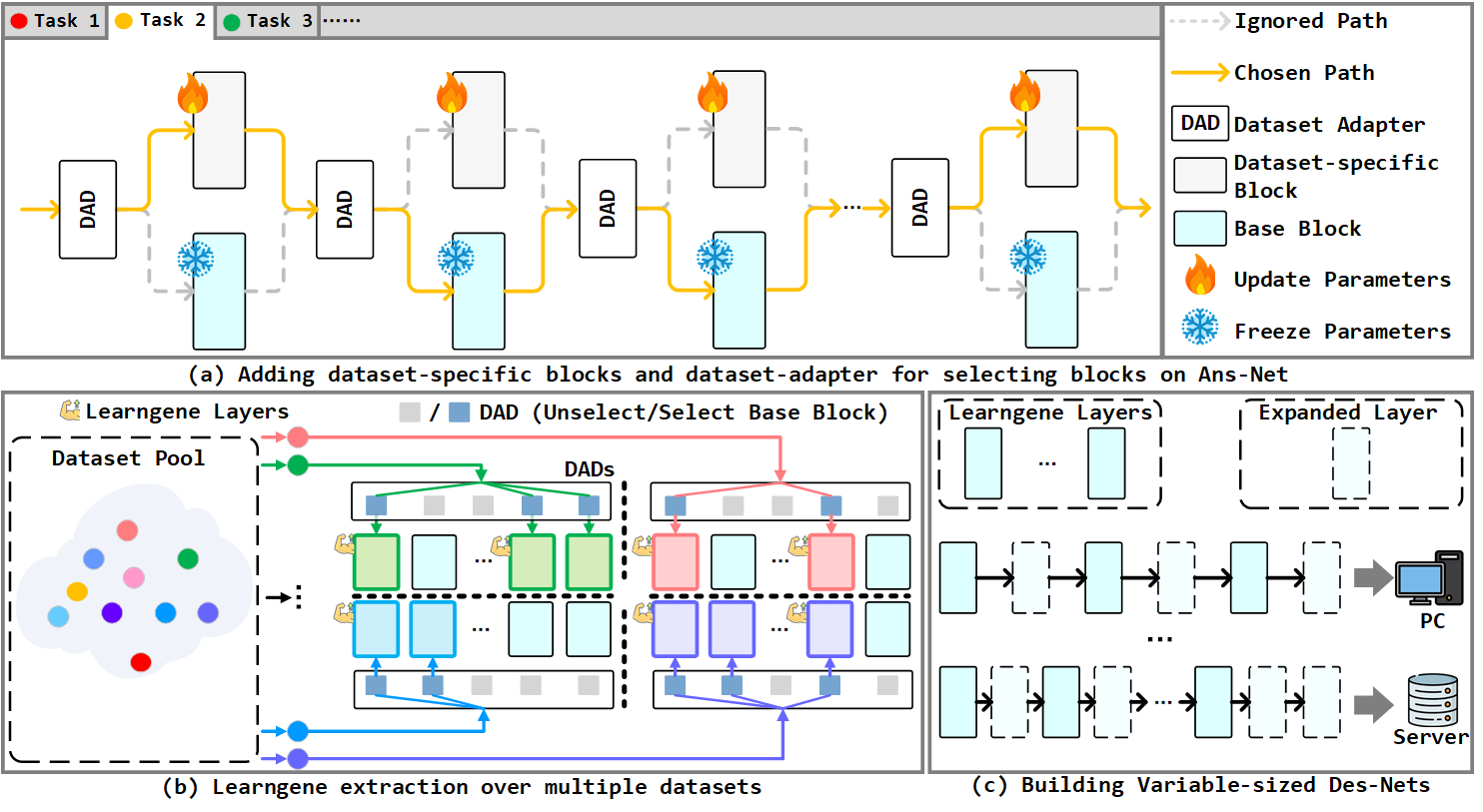}
    \caption[Figure]{The framework of LSAMD. (a) Each layer of the super Ans-Net consists of a dataset-specific block, a base block and a dataset adapter (DAD). (b) Training the Ans-Net on multiple datasets and extract the base blocks which are used by mostly datasets (marked as muscle) as learngene layers based on the chosen of DADs. (c) Variable-sized Des-Nets are built by expanding learngene layers for various resource constraints.}
    \label{fig:structure}
\end{figure*}

\section{Related Work}
% \subsection{Multi-Task Learning}
% Multi-Task Learning (MTL) methods in deep learning share modules across tasks and design multiple output branches to adapt to different tasks. Earlier studies have employed CNN architectures for learning multiple visual tasks \cite{Liu2018EndToEndML,Standley2019WhichTS,Strezoski2019ManyTL,Zamir2020RobustLT}. More recently, Transformer-based MTL methods \cite{Xu2022MTFormerML,Chen2020PreTrainedIP,Mohamed2021SpatioTemporalML,Chen2023ModSquadDM,Bhattacharjee2022MuITAE,Liu2021SwinTH} have shown advantages over CNNs. While MTL typically aims to train a model to perform well on all tasks, this paper focuses on identifying modules consistently used in pre-trained networks across different datasets.
\subsection{Mixture of Experts (MoE)}
The Mixture of Experts (MoE) technique, initially proposed by Jacobs et al. \cite{Jacobs1991AdaptiveMO}, enhances efficiency by combining multiple sub-models and performing conditional computation. Recent research in natural language processing \cite{Shazeer2017OutrageouslyLN,Chen2024textttMoERBenchTB}, vision \cite{Riquelme2021ScalingVW,Wu2022ResidualMO,Liu2024TaskcustomizedMA}, and multimodal \cite{Lin2024MoELLaVAMO,Akbari2023AlternatingGD} applications has adopted sparse MoE techniques to reduce computational costs and train large-scale models. Unlike traditional expert models, modern MoE methods like the Memory Mixture of Experts (MMoE) \cite{Ye2023TaskExpertDA} dynamically assembles task-specific features during multi-task training, significantly enhancing multi-task prediction capabilities. Switch Transformer \cite{fedus2022switch} leveraged distributed training mechanisms and expert sharing techniques. In this paper, we are inspired the MoE method and propose a novel learngene extraction method to decide the propagation path between dataset-related and base blocks.

\subsection{Learngene}
The Learngene involves two key stages: condensing learngene from the Ans-Net, constructing the Des-Net using these learngene and fine-tuning the Des-Net on downstream datasets with minimal steps. The vanilla Learngene (Van-LG) \cite{wang2022learngene} identifies high-level layers as learngene based on the gradient information from Ans-Net during training on multiple tasks. These learngene is combined with a variable number of randomly initialized low-level layers to create Des-Nets of different sizes. Auto-Learngene \cite{Wang2023LearngeneIC} trains pseudo Des-Nets and employs a meta-network to automatically select Ans-Net layers most similar to the pseudo Des-Nets as learngene. Learngene Pool \cite{shi2024building} distills the Ans-Net into multiple smaller models, using their layers as learngene instances to build a pool from which Des-Nets are constructed. TLEG \cite{xia2024transformer} distills Ans-Net into layers that are expanded linearly to create Des-Nets of varying depths. PEG \cite{wangvision} samples Multi-Head Self-Attention and Feed-Forward layers from Ans-Net as learngene using a probabilistic approach, expanding them through non-linear mapping to form Des-Nets. SWS \cite{xia2024exploring} distills Ans-Net into a single model with multiple stages, sharing layer weights, where each stage's layers serve as learngene to guide Des-Net expansion. However, these methods extract learngene from a single dataset, limiting their effectiveness and ultimately impacting the performance of the expanded Des-Nets.

\section{Methodology}
To overcome these limitations, we introduce a new approach called LSAMD, detailed in Figure \ref{fig:structure}. In this section, we outline LSAMD by focusing on two main aspects: searching learngene from the Ans-Net using multiple datasets, and building variable-sized Des-Nets based on the learngene.

\subsection{Learngene Search Across Multiple Datasets}
% \textbf{The choice of the Ans-Net}. 
% To extract learngenes with strong capabilities, we first select an appropriate Ans-Net. In this study, we use DeiT-Base \cite{Touvron2020TrainingDI}, a 12-layer model from the DeiT family, as the Ans-Net. Its larger parameters allow it to acquire extensive knowledge from pre-trained datasets like ImageNet-1K \cite{Russakovsky2014ImageNetLS}, and makes it ideal for extracting effective learngene, which are essential for building Des-Nets.
\textbf{Adding dataset-specific blocks to the Ans-Net}. 
We first employ DeiT-Base \cite{Touvron2020TrainingDI} as the Ans-Net, as its larger number of parameters enables it to acquire extensive knowledge from pre-trained datasets.
Subsequently, we use $M$ datasets, $T_1, T_2, \cdots, T_M$, with varying data distributions to extract learngene. To enable the Ans-Net to learn dataset-specific knowledge while retaining its original capabilities, we introduce an additional component called the dataset-specific block ($h_i^D$), which operates in parallel with the original block called the base block ($h_i^B$) in each layer of the Ans-Net, as illustrated in Figure \ref{fig:structure}a.
Since randomly initialized dataset-specific blocks have limited learning capacity compared to the well-trained base blocks, this setup could disadvantage the dataset-specific blocks during training. To address this, we share the same architecture and initial parameters between the dataset-specific blocks and base blocks in each layer of the Ans-Net.
During training, for each dataset, we update the parameters of the dataset-specific blocks, while keeping the base blocks frozen. This approach ensures that the base blocks remain unaffected by dataset-specific knowledge. Additionally, to prevent interference between the two blocks, we enforce a mechanism that selects only one block per layer for propagation based on the input data.

\textbf{Adding dataset adapter to the Ans-Net}. 
To select between the two blocks, inspired by the Mod-Squad approach \cite{Chen2022ModSquadDM}, we introduce a router-like module called the Dataset Adapter (DAD). The DAD determines the propagation path by selecting between the dataset-specific and base blocks, as shown in Figure \ref{fig:structure}a.
For the $i^{th}$ layer, given input data $x_{t} \in \mathbb{R}^{N \times D}$ from the $t^{th}$ dataset, DAD generates probabilities for each block based on the input: $P(h_i^D|x_t)$ and $P(h_i^B|x_t)$ using the following process:
\begin{equation}
    P(h_i^D|x_t), \ P(h_i^B|x_t) = \operatorname{softmax}(x_tW_1+\operatorname{softplus}(x_tW_{2})),
\end{equation}
where $W_1 \in \mathbb{R}^{D \times 2}$ represents the weight matrices in DAD, and $W_2 \in \mathbb{R}^{D \times 2}$ are noisy weight matrices that enhance DAD's robustness \cite{Shazeer2017OutrageouslyLN}. The $\operatorname{softplus}(\cdot)$ function smooths the ReLU operation:
\begin{equation}
    \operatorname{softplus}(x) = \operatorname{log}(1+e^x).
\end{equation}
In each layer, if more than half of the samples in a batch have a higher probability for a specific block, that block becomes the selected propagation path. For instance, if DAD assigns a higher probability to the base block for more than 50 out of 100 images in a batch, the base block is selected, and the dataset-specific block is ignored.
Alternatively, as Mod-Squad proposes, each sample can independently forward to corresponding block based on the probability, and the outputs from both blocks are concatenated. However, this approach has been found to yield lower performance compared to the method employed here, as discussed in Section \ref{sec:ex_2}. 
After adding the data-specific block and the DAD, the Ans-Net becomes a searchable architecture called the super Ans-Net.

\begin{figure}[t]
    \centering
    \includegraphics[width=0.41\textwidth]{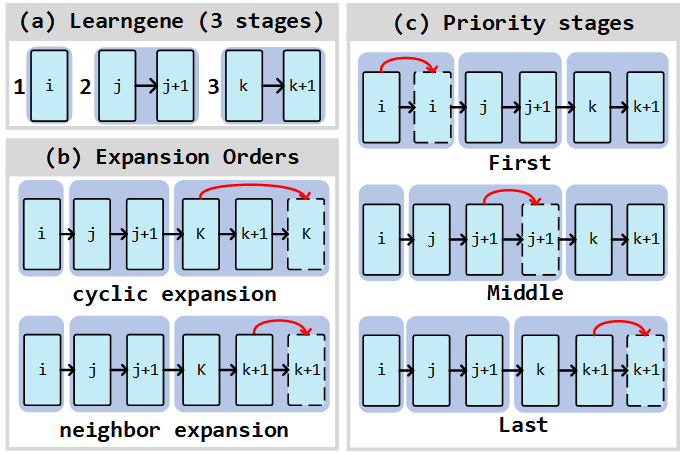}
    \caption[Figure]{(a) Splitting the learngene layers into 3 stages. (b) Two kinds of expansion orders. (c) Three priority expansion stages. $j/j+1$ and $k/k+1$ are adjacent learngene layers. We take 5-layer learngene and 6-layer Des-Net as an example}
    \label{fig:expansion_ways}
\end{figure}

\textbf{Dataset-Adapter for searching paths over multiple datasets}.
Then, we train the super Ans-Net on multiple datasets, as illustrated in Figure \ref{fig:structure}b. During training, DAD searches between dataset-specific and base blocks in each layer, and DAD returns the corresponding search path based on the input dataset at the end of training. 
In these paths, we calculate the usage of the searched base blocks by comparing $P(h_i^D|x_t)$ and $P(h_i^B|x_t)$:
\begin{equation}
\begin{aligned}
    \operatorname{G_i} & = \sum_{t=1}^{M}g_i(t), \\ 
    \text{s.t.} \quad g_i(t) & =\left\{
\begin{aligned}
1, & \quad \text{if} \ P(h_i^B|x_t) \geq P(h_i^D|x_t) \\
0, & \quad \text{otherwise}
\end{aligned}
\right.
\end{aligned}
\end{equation}
Here, $g_i(t) \in \{0, 1\}$ indicates whether the $i^{th}$ base block is used on the $t^{th}$ dataset, and $\operatorname{G_i}$ represents the total usage of the $i^{th}$ base block across all datasets.
We then divide the searched base blocks into two scenarios: heavily used and lightly used, depending on whether they are used by more than $\tau \in \mathbb{Z} \cap [1, M]$ datasets.
In this paper, we set $\tau$ to $M/2$ for ensuring that base blocks used by more than half of the datasets, which helps identify blocks from the Ans-Net that are most consistently shared across diverse datasets.
Based on the generality of these heavily used blocks, we extract them as "learngene" layers. For example, as shown in Figure \ref{fig:structure}b, the first and third base blocks are selected as learngene layers because they are utilized by more than half of the datasets. In this study, we take 12 datasets (10 classification datasets and 2 segmentation datasets) for diversity. Here, we set $\tau = 6$, which corresponds to half the total number of 10 datasets, and the extracted learngene layers are indexed at 1, 4, 5, 7, and 8.

\begin{figure*}[t]
    \centering
    \begin{subfigure}[b]{0.235\textwidth}
        \includegraphics[width=\textwidth]{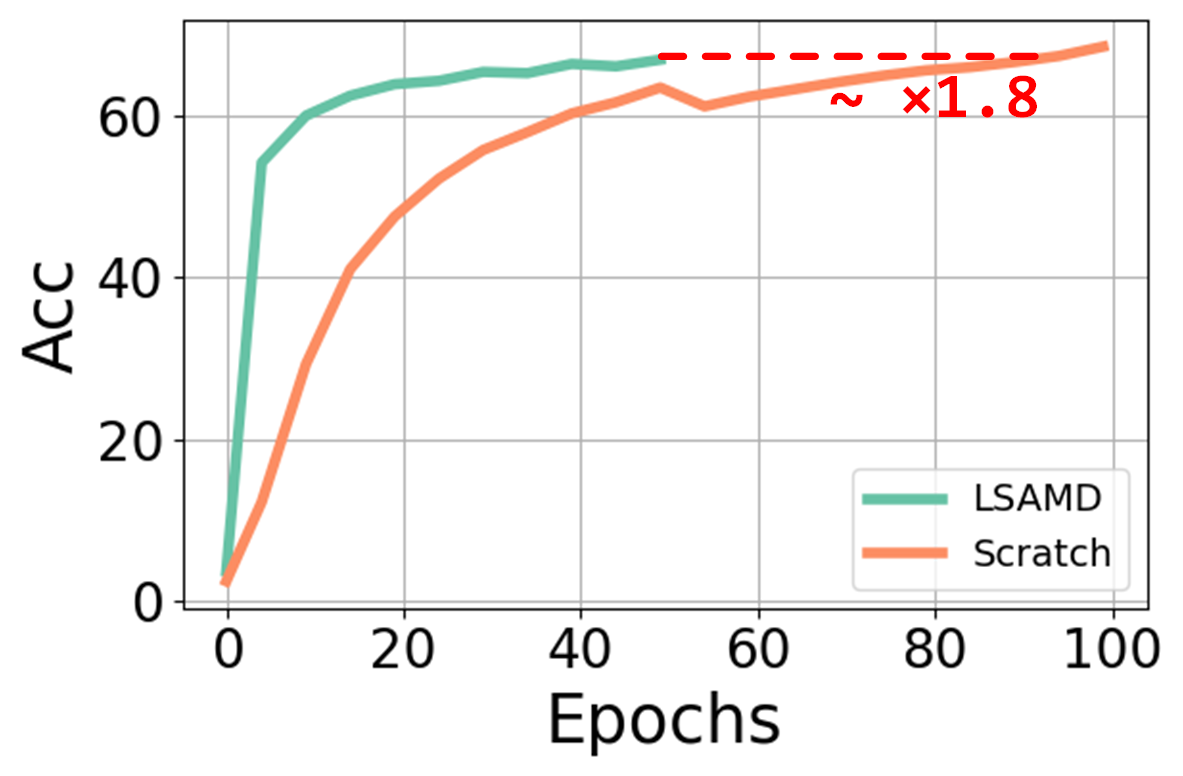}
        \caption{6-layer Des-Net/IMNet-1k}
        \label{fig_ours_pf_imnet_6L}
    \end{subfigure}
    \begin{subfigure}[b]{0.235\textwidth}
        \includegraphics[width=\textwidth]{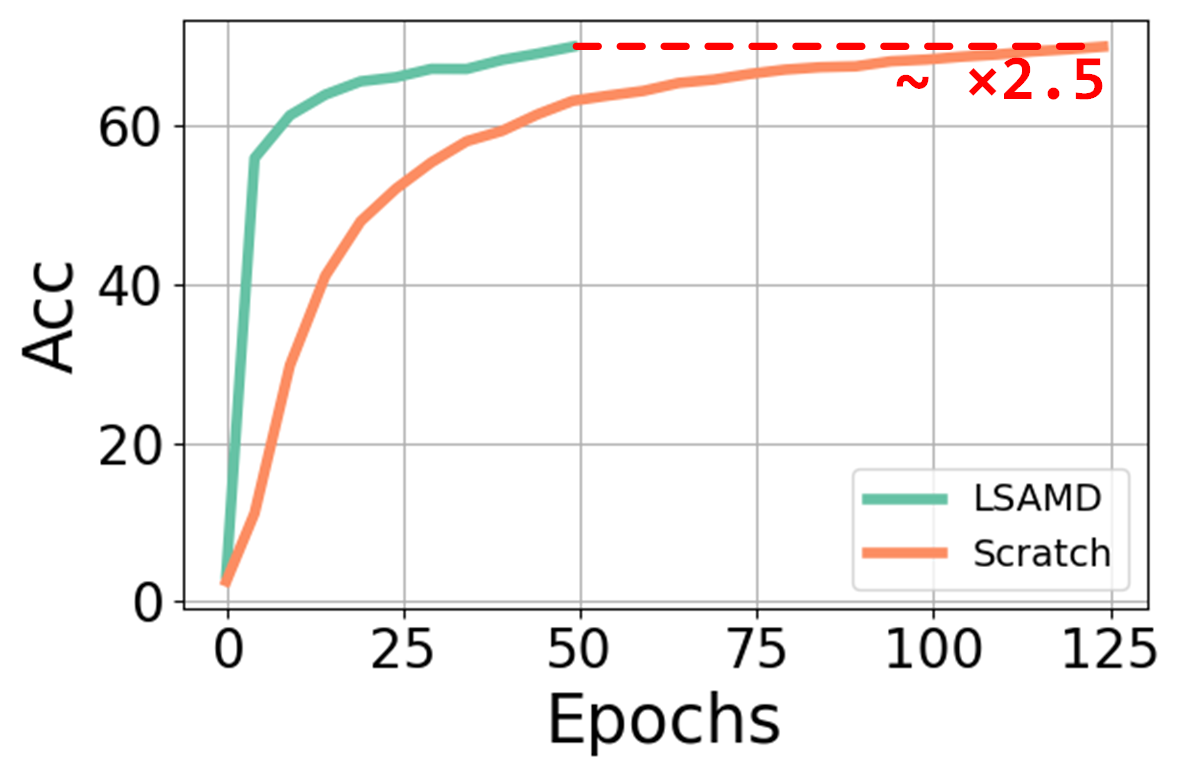}
        \caption{7-layer Des-Net/IMNet-1k}
        \label{fig_ours_pf_imnet_7L}
    \end{subfigure}
    \begin{subfigure}[b]{0.235\textwidth}
        \includegraphics[width=\textwidth]{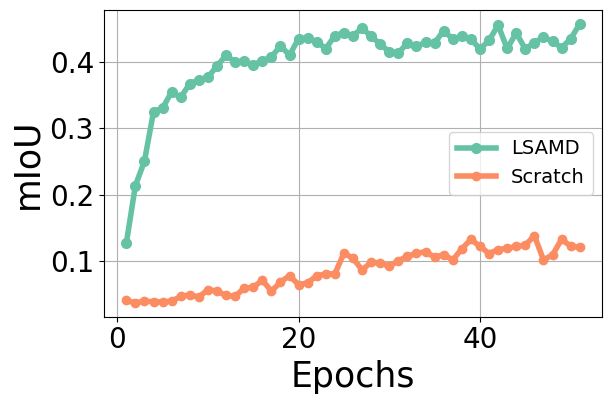}
        \caption{6-layer Des-Net/ADE-20K}
        \label{fig_ours_pf_ade_6L}
    \end{subfigure}
    \begin{subfigure}[b]{0.235\textwidth}
        \includegraphics[width=\textwidth]{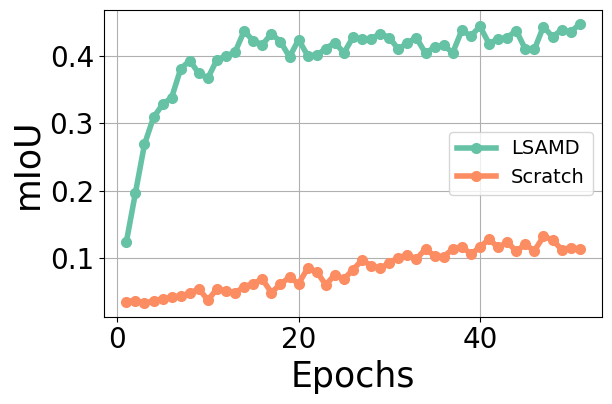}
        \caption{7-layer Des-Net/ADE-20K}
        \label{fig_ours_pf_ade_7L}
    \end{subfigure}
    \caption{Comparison of Scratch and LSAMD on Des-Nets with 6 and 7 layers on IMNet-1K and ADE-20K datasets.}
    \label{fig_ours_pf_imnet}
\end{figure*}

While our method, LSAMD, is inspired by the routing mechanisms in Mixture-of-Experts (MoE), its fundamental goal and implementation are distinct. MoE aims to accelerate inference by sparsely activating task-specific experts. In contrast, LSAMD follows the Learngene paradigm to extract a core set of foundational components from a large Ans-Net for efficiently initializing smaller, resource-adaptive Des-Nets.
This distinction in purpose leads to three key differences. First, in terms of granularity, LSAMD performs coarse-grained routing between entire blocks to identify a set of reusable layers, whereas MoE conducts fine-grained routing among MLPs within a block. Second, regarding scope, LSAMD seeks foundational blocks reused across datasets, while MoE selects experts per-dataset for specialized execution. Finally, in architectural impact, LSAMD reduces the model's core depth to enable flexible scaling, while MoE retains the full model depth, only reducing the activation width during inference.
% Although LSAMD shares a conceptual similarity with Mod-Squad, there are significant differences between the two approaches. 
% First, Mod-Squad is designed to train a network that performs well across multiple tasks, whereas LSAMD focuses on search a proper propagation path for each dataset in pre-trained models. 
% Second, Mod-Squad selects a subset of MLP layers at the element level for propagation through a routing network, while LSAMD operates at the block level, selecting one of two blocks in each layer via the DAD module.
% Finally, Mod-Squad ranks experts within each task based on each image, whereas LSAMD determines block selection based on a batch of data.

\subsection{Building Variable-Sized Des-Nets}
After extracting learngene layers, we construct variable-sized Des-Nets by replicating and adding these layers, as illustrated in Figure \ref{fig:structure}c. Since learngene layers retain both critical knowledge and positional information, we preserve their original order from the Ans-Net and organize them into stages, where adjacent learngene layers form a single stage, as shown in Figure \ref{fig:expansion_ways}a.
Within stages containing multiple blocks, there are two expansion options: cyclic expansion and neighbor expansion. Cyclic expansion replicates learngene layers in their original sequence within the stage, as depicted in Figure \ref{fig:expansion_ways}b top. Neighbor expansion duplicates the last learngene layer of the stage, as shown in Figure \ref{fig:expansion_ways}b bottom. Given that neighbor expansion better preserves the original order and learning process of the Ans-Net, we adopt this method for expansion.

Furthermore, it is also crucial to prioritize which stages of the learngene layers to expand. For Des-Nets with the same number of layers, we consider three expansion priorities: the first, middle, or last stage, as illustrated in Figure \ref{fig:expansion_ways}c. Our investigation, detailed in Section \ref{sec:ex_2}, shows that expanding in the last-middle-first priority order yields the best results for the Des-Nets, and we take this priority in our methods.

\begin{table}[t]
\centering
\begin{tabular}{c|c|c|c|c}
\toprule[1pt]
Layer            & Method & Cifar100 & Cifar10 & Pets  \\ \midrule
\multirow{3}{*}{6}  & SWS   &80.39    &94.87  & 75.94          \\  
                  & TLEG  &77.01     &94.27    &  64.44         \\  
                & SLAMD   &\textbf{82.77}   &\textbf{96.67}    &\textbf{79.86}  \\ \midrule
\multirow{3}{*}{7} & SWS   &82.19     &95.99     & 81.09         \\  
                  & TLEG  &80.03     &95.81    & 74.60         \\ 
                 & SLAMD   &\textbf{84.38}     &\textbf{97.42}    &\textbf{81.34}  \\ \midrule
\multirow{3}{*}{8}  & SWS   &83.61     &96.54        &80.25          \\
                 & TLEG  &81.40     &96.43    & 76.35         \\  
                 & SLAMD   &\textbf{83.78}     &\textbf{96.74}    &\textbf{80.90} \\ \bottomrule[1pt]
\end{tabular}
\caption{Comparisons between LSAMD and SWS, TLEG on 4 downstream datasets of Des-Nets with various layer.}
\centering 
\label{tab:xia_ours}
\end{table}

\begin{figure}[t]
    \centering
    \begin{subfigure}[b]{0.23\textwidth}
        \includegraphics[width=\textwidth]{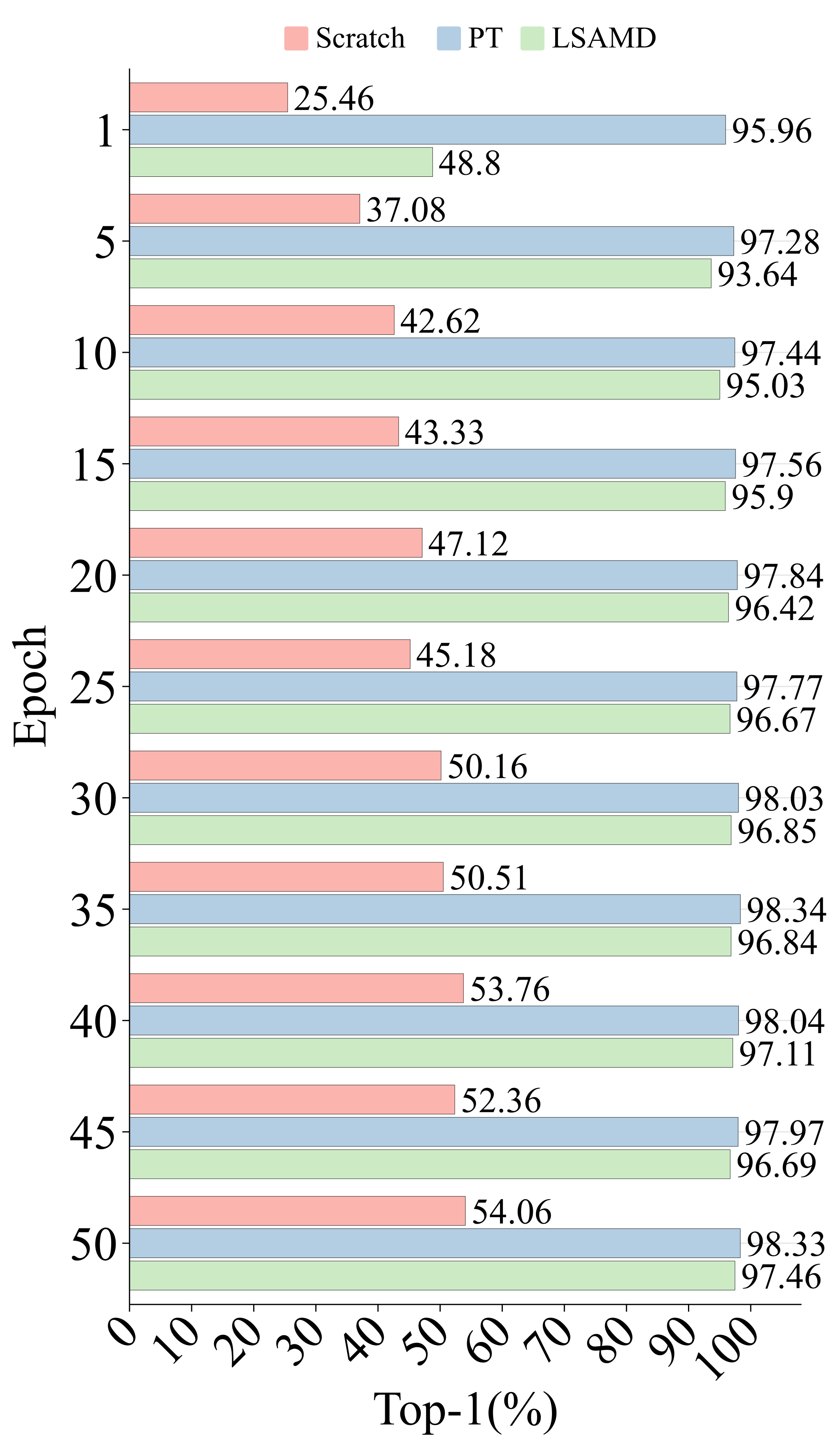}
        \caption{Cifar10}
    \end{subfigure}
    \begin{subfigure}[b]{0.23\textwidth}
        \includegraphics[width=\textwidth]{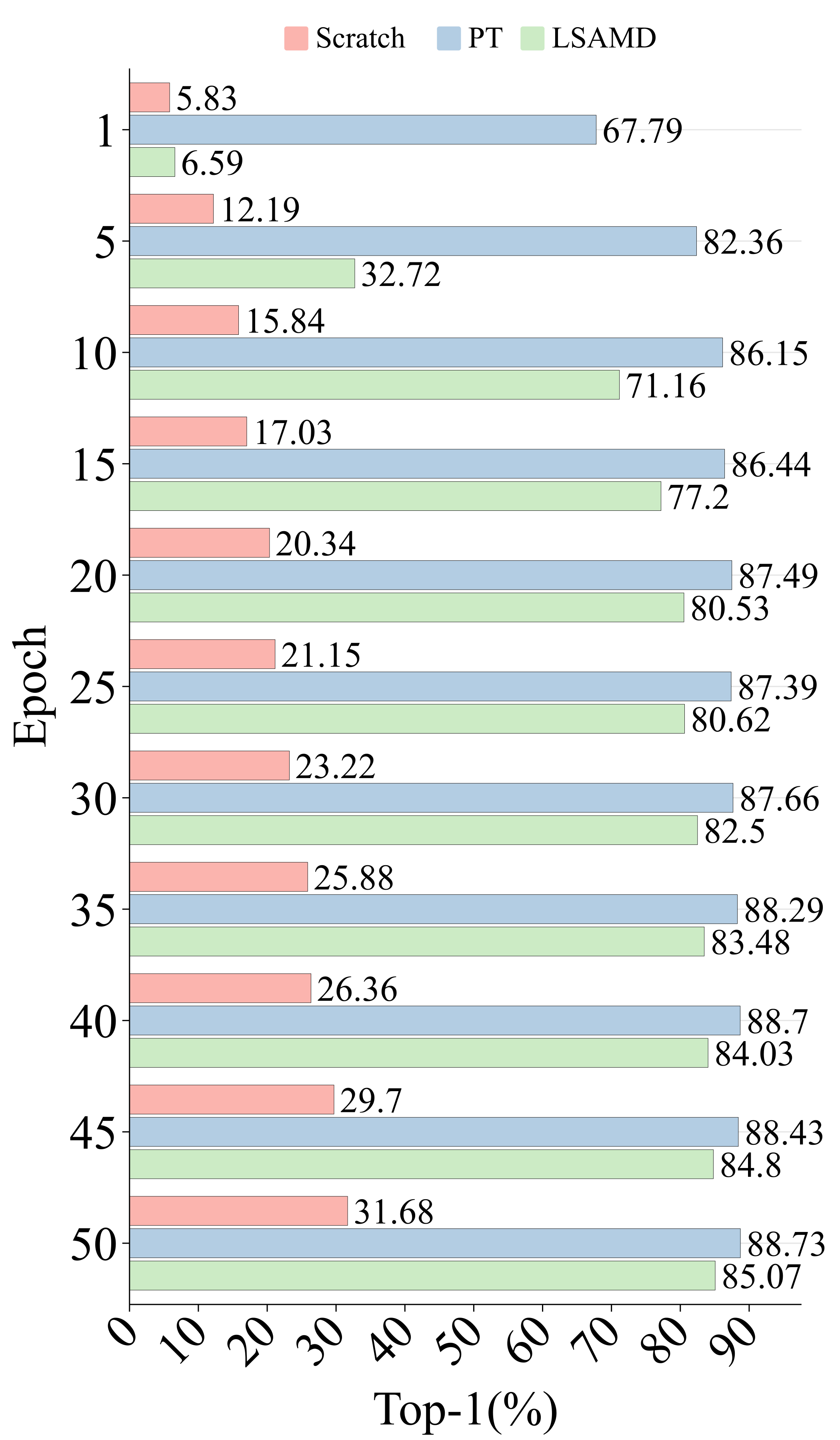}
        \caption{Cifar100}
    \end{subfigure}
    \caption{Performance comparison of Scratch, Pretrain-Finetune (PF) and LSAMD on 10-layer Des-Nets across 4 downstream image classification dataset.}
    \label{fig_ours_pf_downstream}
\end{figure}

\section{Experiments}
\label{sec:ex}

\subsection{Experimental Setup}
We train the super Ans-Net by the following 12 datasets: Cifar100 \cite{Krizhevsky2009LearningML}, Cifar10 \cite{Krizhevsky2009LearningML}, Tiny-ImageNet (Ti-IMNet) \cite{Le2015TinyIV}, INAT2019 \cite{Horn2017TheIS}, Food101 \cite{Bossard2014Food101M}, Cars196 \cite{Krause20133DOR}, Flowers102 \cite{Nilsback2008AutomatedFC}, Pets \cite{Parkhi2012CatsAD}, DTD \cite{Cimpoi2013DescribingTI}, Caltech101 \cite{FeiFei2004LearningGV}, VOC12 \cite{everingham2015pascal}, and ADE-20K \cite{zhou2019semantic,zhou2017scene}. The overview information of these ten datasets and the training details are shown in the supplementary material.
% The Ans-Net is trained for 50 epochs on each dataset, during which we record the DAD selections for learngene extraction. The resulting Des-Nets are then fine-tuned on downstream datasets for 50 epochs. These downstream datasets are sampled from the ten datasets mentioned previously. Since the parameters of the learngene layers are frozen during extraction, they do not incorporate knowledge from the downstream datasets, ensuring a fair comparison of Des-Net performance on these datasets. Another hyper-parameters are listed in the supplementary material. 
% We evaluate Des-Net performance primarily using Top-1 classification accuracy (Top-1(\%)). Additionally, we report the number of parameters (Params) and FLOPs to describe the size and computational complexity of the Des-Nets.

\begin{figure*}[t]
    \centering
    \begin{subfigure}[b]{0.23\textwidth}
        \includegraphics[width=\textwidth]{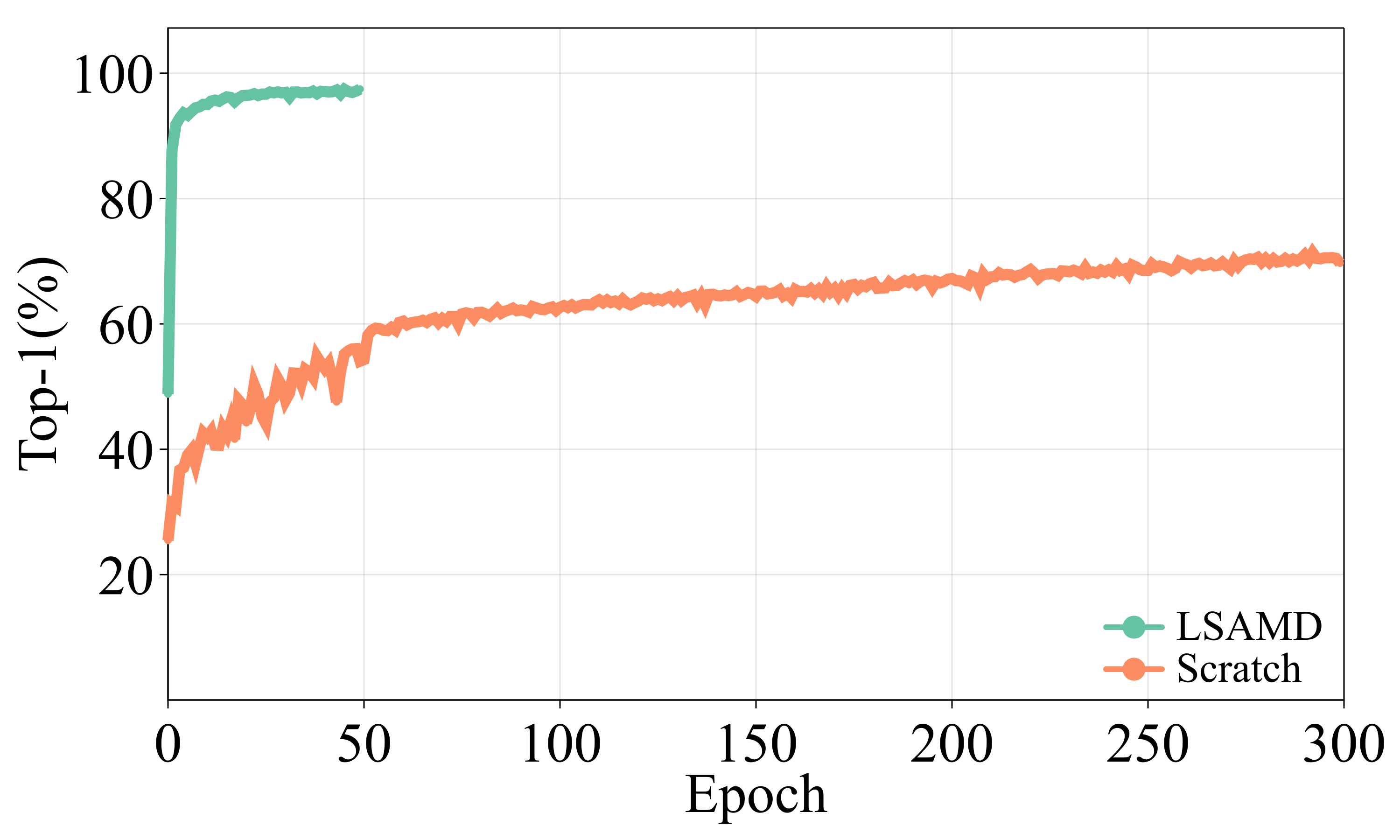}
        \caption{Cifar10}
        \label{fig:image1}
    \end{subfigure}
    \begin{subfigure}[b]{0.23\textwidth}
        \includegraphics[width=\textwidth]{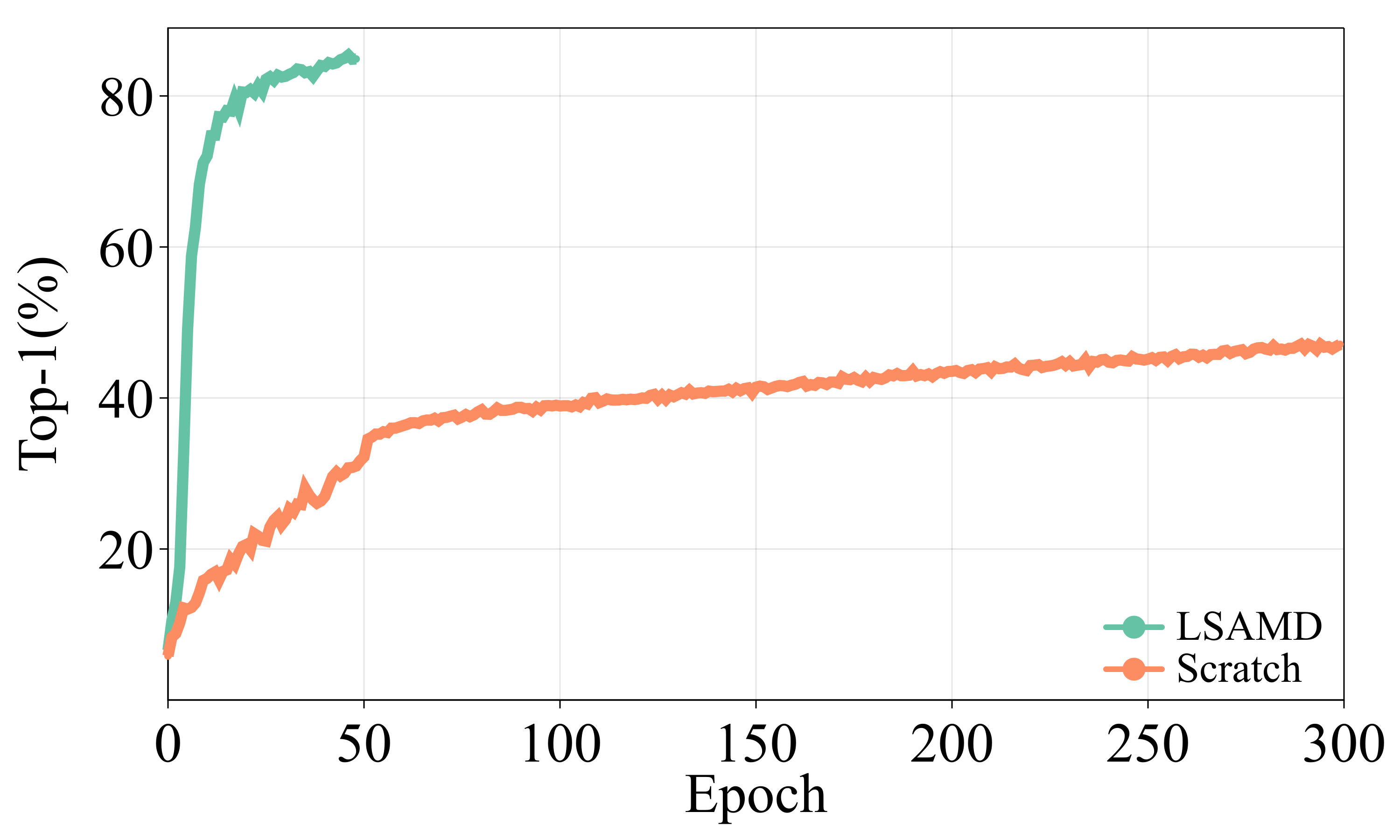}
        \caption{Cifar100}
        \label{fig:image2}
    \end{subfigure}
    \begin{subfigure}[b]{0.23\textwidth}
        \includegraphics[width=\textwidth]{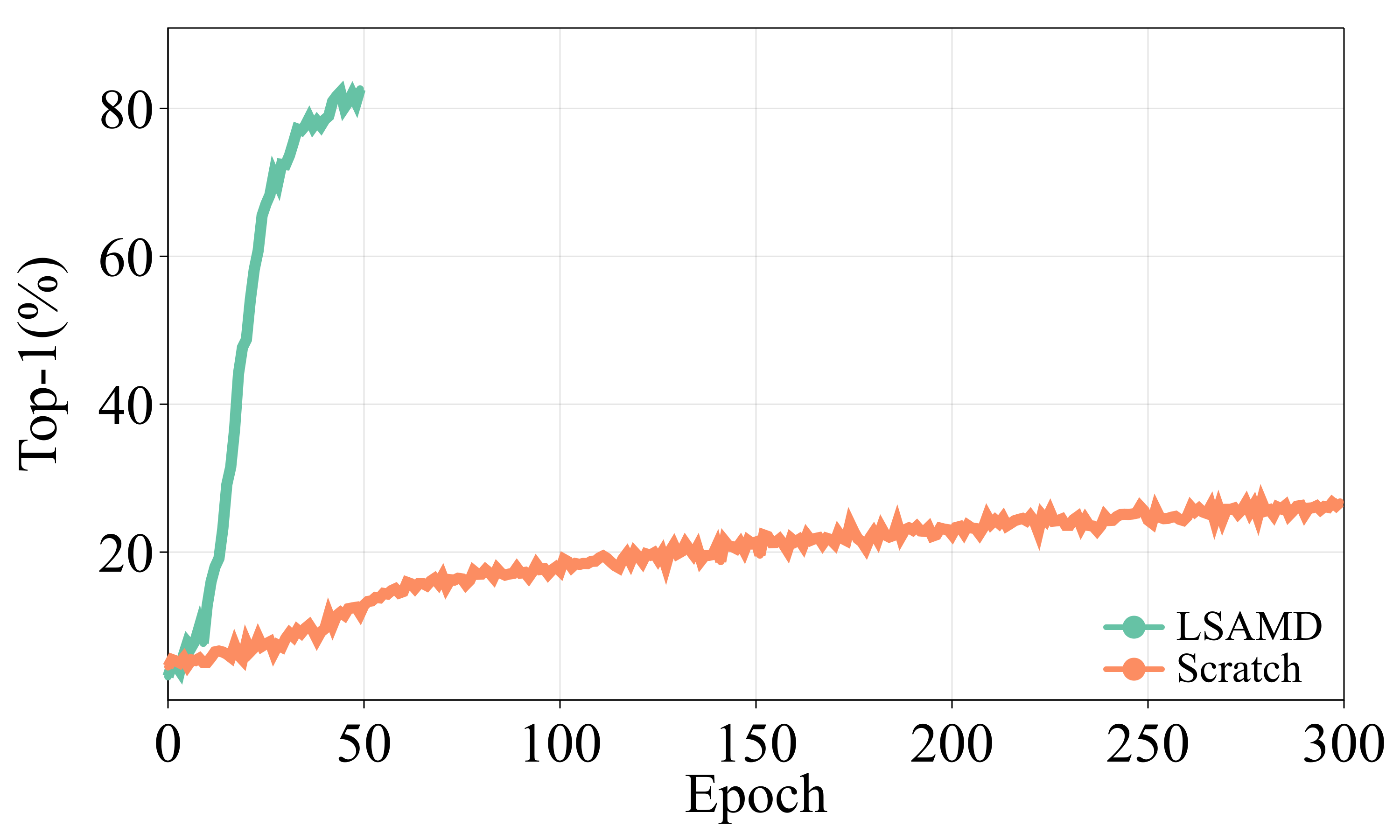}
        \caption{Pets}
        \label{fig:image3}
    \end{subfigure}
    \begin{subfigure}[b]{0.23\textwidth}
        \includegraphics[width=\textwidth]{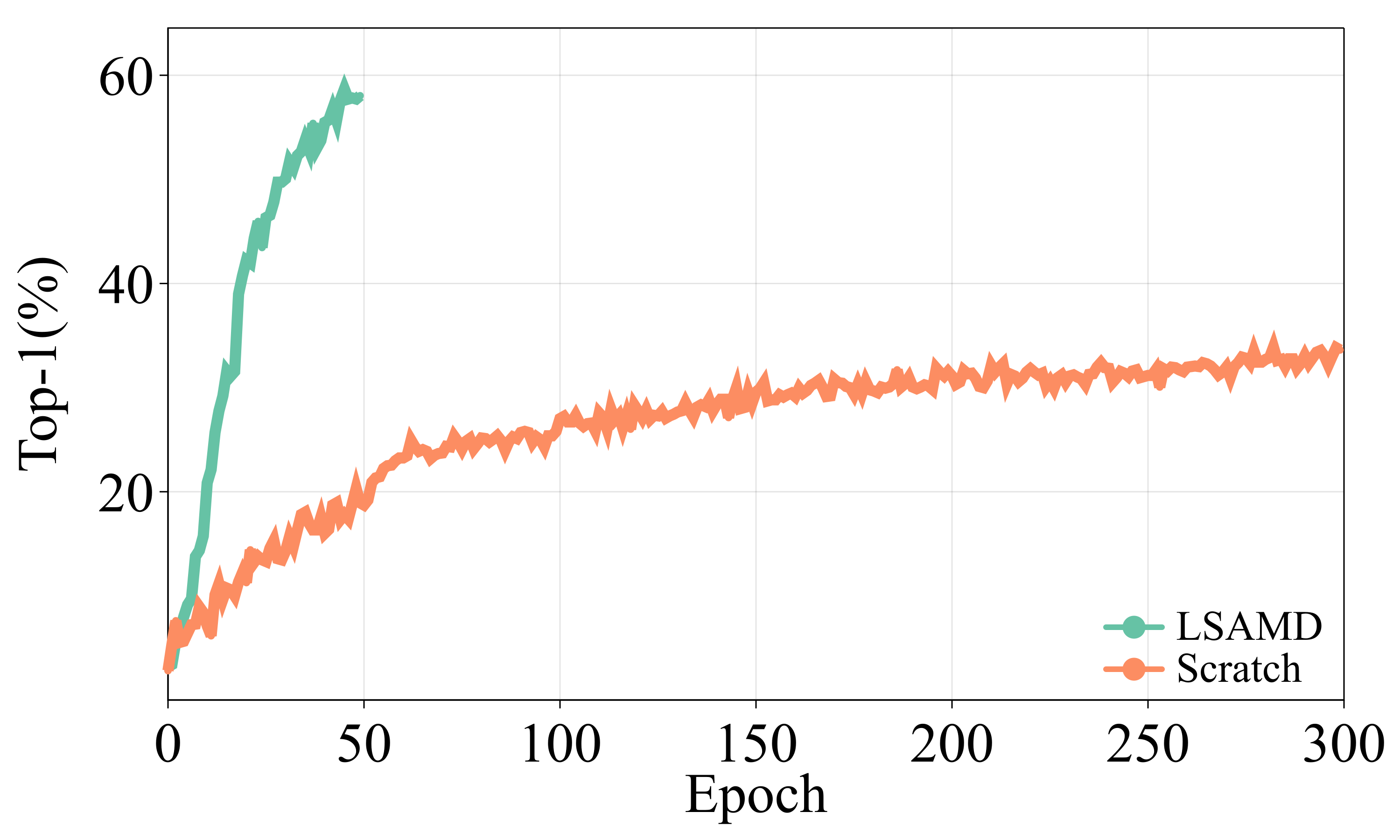}
        \caption{DTD}
        \label{fig:image3}
    \end{subfigure}
    \caption{Comparison of Scratch and LSAMD on 10-layer Des-Nets across 4 downstream dataset.}
    \label{fig_ours_scratch_downstream}
\end{figure*}

% \usepackage{multirow}
\begin{table*}[]
\centering
\begin{tabular}{c|c|c|ccccccc}
\toprule[1pt]
\multirow{2}{*}{Layer} & \multirow{2}{*}{Params(M)} & \multirow{2}{*}{FLOPs(G)} & \multicolumn{7}{c}{LSAMD}                                                                                                                                                                                                \\ 
                       &                            &                           & \multicolumn{1}{c|}{LG-Params(M)}            & \multicolumn{1}{c|}{Cifar100} & \multicolumn{1}{c|}{Cifar10} & \multicolumn{1}{c|}{Tiny-IMNet} & \multicolumn{1}{c|}{INAT-2019} & \multicolumn{1}{c|}{Food-101} & DTD     \\ \midrule
5                      & 36.11                      & 7.09                      & \multicolumn{1}{c|}{\multirow{11}{*}{36.11}} & \multicolumn{1}{c|}{80.93}  & \multicolumn{1}{c|}{95.70} & \multicolumn{1}{c|}{71.53}    & \multicolumn{1}{c|}{50.26}   & \multicolumn{1}{c|}{58.68}  & 52.39 \\
6                      & 43.20                      & 8.49                      & \multicolumn{1}{c|}{}                        & \multicolumn{1}{c|}{81.24}  & \multicolumn{1}{c|}{95.94} & \multicolumn{1}{c|}{70.97}    & \multicolumn{1}{c|}{53.49}   & \multicolumn{1}{c|}{57.08}  & 53.62 \\ 
7                      & 50.28                      & 9.89                      & \multicolumn{1}{c|}{}                        & \multicolumn{1}{c|}{82.25}  & \multicolumn{1}{c|}{96.59} & \multicolumn{1}{c|}{72.39}    & \multicolumn{1}{c|}{54.26}   & \multicolumn{1}{c|}{59.17}  & 54.20 \\
8                      & 57.37                      & 11.28                     & \multicolumn{1}{c|}{}                        & \multicolumn{1}{c|}{83.78}  & \multicolumn{1}{c|}{96.74} & \multicolumn{1}{c|}{73.27}    & \multicolumn{1}{c|}{56.57}   & \multicolumn{1}{c|}{63.81}  & 58.56 \\ 
9                      & 64.46                      & 12.68                     & \multicolumn{1}{c|}{}                        & \multicolumn{1}{c|}{84.63}  & \multicolumn{1}{c|}{97.30} & \multicolumn{1}{c|}{74.63}    & \multicolumn{1}{c|}{58.92}   & \multicolumn{1}{c|}{65.96}  & 59.47 \\ 
10                     & 71.55                      & 14.07                     & \multicolumn{1}{c|}{}                        & \multicolumn{1}{c|}{85.40}  & \multicolumn{1}{c|}{97.46} & \multicolumn{1}{c|}{76.40}    & \multicolumn{1}{c|}{61.85}   & \multicolumn{1}{c|}{67.35}  & 58.67 \\
11                     & 78.64                      & 15.47                     & \multicolumn{1}{c|}{}                        & \multicolumn{1}{c|}{85.52}  & \multicolumn{1}{c|}{97.59} & \multicolumn{1}{c|}{75.98}    & \multicolumn{1}{c|}{64.72}   & \multicolumn{1}{c|}{64.36}  & 55.69 \\ 
12                     & 85.72                      & 16.86                     & \multicolumn{1}{c|}{}                        & \multicolumn{1}{c|}{83.86}  & \multicolumn{1}{c|}{97.70} & \multicolumn{1}{c|}{76.62}    & \multicolumn{1}{c|}{65.70}   & \multicolumn{1}{c|}{64.21}  & 57.61 \\ 
13                     & 92.81                      & 18.26                     & \multicolumn{1}{c|}{}                        & \multicolumn{1}{c|}{77.64}  & \multicolumn{1}{c|}{97.70} & \multicolumn{1}{c|}{77.06}    & \multicolumn{1}{c|}{64.74}   & \multicolumn{1}{c|}{60.71}  & 55.53 \\
14                     & 99.90                      & 19.65                     & \multicolumn{1}{c|}{}                        & \multicolumn{1}{c|}{86.18}  & \multicolumn{1}{c|}{97.65} & \multicolumn{1}{c|}{77.56}    & \multicolumn{1}{c|}{65.52}   & \multicolumn{1}{c|}{62.29}  & 56.78 \\  
15                     & 106.99                     & 21.05                     & \multicolumn{1}{c|}{}                        & \multicolumn{1}{c|}{85.77}  & \multicolumn{1}{c|}{97.97} & \multicolumn{1}{c|}{77.41}    & \multicolumn{1}{c|}{66.28}   & \multicolumn{1}{c|}{65.72}  & 57.07 \\ 
\bottomrule[1pt]
\end{tabular}
\caption{The results of variable-sized Des-Nets on multiple downstream image classification datasets and comparison of storage resource costs between training from scratch and LSAMD. `Layer' means the number of layers in the Des-Nets, and `LG-Params' is number of parameters of the extracted learngene.}
\label{ex_downstream_datasets}
\end{table*}

\subsection{Main Results and Analysis}
\textbf{LSAMD vs. Scratch}. 
We compare the performance of LSAMD and training from scratch (Scratch) on 6-layer and 7-layer Des-Nets on the ImageNet-1K (IMNet) \cite{Deng2009ImageNetAL} dataset, as shown in Figure \ref{fig_ours_pf_imnet}. It can be observed that LSAMD achieves superior performance on both 6-layer and 7-layer Des-Nets and both tasks. 
Specifically, when training a 6-layer Des-Net, LSAMD reduces training costs by approximately \textbf{1.8×} while reaching comparable performance, as illustrated in Figure \ref{fig_ours_pf_imnet_6L}. 
Similarly, for the 7-layer Des-Net, LSAMD achieves a cost reduction of around \textbf{2.5×}, as depicted in Figure \ref{fig_ours_pf_imnet_7L}. These results highlight the advantage of LSAMD over the Scratch method on the ImageNet-1K dataset.
Moreover, as the number of layers in Des-Net increases, the training cost savings become more significant. This trend demonstrates the scalability of LSAMD and its suitability for scenarios with fewer resource constraints. 
Moreover, we also compare LSAMD and training from scratch on the semantic segmentation task. As shown in Figure~\ref{fig_ours_pf_ade_6L} and Figure~\ref{fig_ours_pf_ade_7L}, both Des-Nets initialized by LSAMD maintain excellent results compared to the training from scratch.

Furthermore, We train the 10-layer Des-Net from scratch (Scratch) for 300 epochs and compare it with the same Des-Net initialized using LSAMD, which is trained for only 50 epochs on Cifar10, Cifar100, Pets and DTD. 
As illustrated in Figure \ref{fig_ours_scratch_downstream}, the LSAMD-initialized Des-Net significantly outperforms the Scratch across all datasets. 
Moreover, even after 300 epochs of training, the Scratch shows limited improvement on downstream datasets with smaller sample sizes. 
In contrast, the LSAMD-initialized Des-Nets achieve superior performance after 50 epochs, highlighting its ability to enhance performance on datasets with fewer samples. 
These results underscore the advantages of LSAMD over training from scratch on both classification and segmentation tasks.

\textbf{LSAMD vs. PF}. We also compare LSAMD against the pretrain-finetune (PF) method, which \textbf{serves as an upper bound}. PF trains the Des-Net on ImageNet-1K for 100 epochs and then finetunes the Des-Net on downstream datasets for 50 epochs. \textbf{In contrast, LSAMD directly trains the Des-Net for 50 epochs.}
Figure \ref{fig_ours_pf_downstream} presents a performance comparison of the Des-Net trained using PF, Scratch, and LSAMD over 50 epochs, with evaluations recorded every 5 epochs. 
The performance of LSAMD is significantly lower than that of PF at the beginning. However, as training progresses, the performance of LSAMD gradually approaches that of PF and ultimately achieves comparable results by the end of training. 
For example, on the Cifar10 dataset, the performance gap between LSAMD and PF decreases from 47.16\% at the start of training to 0.87\% by the end. Similarly, on the Cifar100 dataset, the gap reduces from 61.2\% to 3.66\%. 
These findings demonstrate that LSAMD effectively enhances the training efficiency and performance of Des-Net.

\begin{table*}[]
\begin{tabular}{c|c|c|c|c|c|c|c|c|c|c}
\toprule[1pt]
 & Cifar100 & Cifar10 & Ti-IMNet & INAT2019 & Food101 & Cars196 & Flowers102 & Pets & DTD   & Caltech101 \\ \midrule
Batch   & \textbf{90.71}    & \textbf{98.61}   & \textbf{91.41}         & 58.21     &\textbf{ 80.92}    & \textbf{86.41}    & \textbf{94.20}      & \textbf{85.29}           & \textbf{75.43} & \textbf{99.65}      \\ \midrule
Img     & 87.13    & 95.29   & 84.28         & \textbf{59.87}     & 61.14    & 55.45    & 91.27       & 83.16           & 70.48 & 97.12      \\
\bottomrule[1pt]
\end{tabular}
\caption{Performance comparison of two forward propagation methods for input batch data. `Batch‘ means that entire batch of data selects one block for forward propagation, and `Img' denotes that each sample in one batch independently forwards to the corresponding blocks according to the probability.}
\centering 
\label{tab:propagation_ways}
\end{table*}

\begin{table}[]
\resizebox{0.5\textwidth}{!}{
\begin{tabular}{c|ccc|ccc}
\toprule
\multirow{2}{*}{Dataset} & \multicolumn{3}{c|}{Des-Net with 6 layers}                       & \multicolumn{3}{c}{Des-Net with 7 layers}                       \\ 
                         & \multicolumn{1}{c|}{First} & \multicolumn{1}{c|}{Middle} & Last  & \multicolumn{1}{c|}{First} & \multicolumn{1}{c|}{Middle} & Last  \\ \midrule
Cifar100                 & \multicolumn{1}{c|}{81.24} & \multicolumn{1}{c|}{82.29}  & \textbf{82.77} & \multicolumn{1}{c|}{83.09} & \multicolumn{1}{c|}{83.46}  & \textbf{84.38} \\ 
Cifar10                  & \multicolumn{1}{c|}{95.94} & \multicolumn{1}{c|}{96.00}     & \textbf{96.67} & \multicolumn{1}{c|}{96.65} & \multicolumn{1}{c|}{97.00}     & \textbf{97.42} \\ 
Ti-IMNet                 & \multicolumn{1}{c|}{70.97} & \multicolumn{1}{c|}{72.01}  & \textbf{73.59} & \multicolumn{1}{c|}{72.66} & \multicolumn{1}{c|}{72.39}  & \textbf{74.70}  \\ 
Food101                  & \multicolumn{1}{c|}{57.08} & \multicolumn{1}{c|}{59.73}  & \textbf{63.27} & \multicolumn{1}{c|}{58.38} & \multicolumn{1}{c|}{61.11}  & \textbf{65.64} \\ 
Pets                     & \multicolumn{1}{c|}{67.33} & \multicolumn{1}{c|}{75.23}  & \textbf{79.86} & \multicolumn{1}{c|}{72.07} & \multicolumn{1}{c|}{76.43}  & \textbf{81.34} \\ 
DTD                      & \multicolumn{1}{c|}{53.62} & \multicolumn{1}{c|}{\textbf{55.37}}  & 54.84 & \multicolumn{1}{c|}{53.72} & \multicolumn{1}{c|}{53.94}  & \textbf{56.91} \\ 
Caltech101               & \multicolumn{1}{c|}{43.55} & \multicolumn{1}{c|}{48.68}  & \textbf{50.46} & \multicolumn{1}{c|}{45.05} & \multicolumn{1}{c|}{47.35}  & \textbf{53.80}  \\ \bottomrule
\end{tabular}}
\caption{Comparison of the Des-Net with 6 and 7 layers expanded by three priority expansion stages. `First'/`Middle'/`Last' are first/middle/last-stage priority.}
\label{lab_priority}
\end{table}

\textbf{LSAMD vs. SWS/TLEG}.
To evaluate the effectiveness of extracting learngene using multiple datasets versus a single dataset, we compared LSAMD with a part of earlier Learngene methods: SWS and TLEG for the following reasons. While Van-LG and Auto-Learngene are designed for CNN or ViT architectures, LSAMD is conducted on the DeiT architecture. Additionally, PEG and Learngene Pool report results for DeiT-Ti/S on the downstream datasets, whereas LSAMD primarily focuses on DeiT-B. Therefore, SWS and TLEG were chosen for comparison. For consistency, we compared LEG with SWS and TLEG under the same settings.
Table \ref{tab:xia_ours} shows results on four downstream datasets of 6, 7, 8-layer Des-Nets initialized by LSAMD and SWS/TLEG. LSAMD outperforms all other methods on all datasets. For example, LSAMD exceeds TLEG by 4.23\% on the Cifar100 dataset with a 6-layer Des-Net. 
Although SWS and TLEG use the larger ImageNet-1K dataset to extract learngene, using multiple smaller datasets outperforms them on most downstream datasets. \textbf{This result confirms that extracting learngene by multiple datasets yields better performance than using a single large dataset.}

\textbf{Comparison of storage and training costs between LSAMD and PF}. 
As mentioned above, the Pretrain-Finetune (PF) method is both time-consuming and storage-demanding. In this section, we provide a detailed comparison of the training time and storage costs between LSAMD and PF.
Specifically, LSAMD requires fewer training epochs than PF. For example, when applied to five Des-Nets across a single downstream dataset, LSAMD reduces training costs by approximately \textbf{1.36×} compared to PF (50$\times$10+50$\times$5 epochs vs. 150$\times$5 epochs). It is important to note that the model is pretrained for 100 epochs and fine-tuned for 50 epochs per dataset, and the cost savings become more pronounced as the scale of Des-Nets increases.
Furthermore, we compare the storage requirements of PF and LSAMD, as shown in Table \ref{ex_downstream_datasets}. LSAMD achieves a substantial reduction in storage costs, saving \textbf{95.41\%} of the storage space required by PF (36.11M vs. 787.03M). This significant saving is attributed to the fact that PF stores all parameters of variable-sized pre-trained models, whereas LSAMD only stores the learngene.

\textbf{Performance of variable-sized Des-Nets}. 
We start from learngene layers and use them to initialize Des-Nets with different scales. 
Table \ref{ex_downstream_datasets} shows the performance of these variable-sized Des-Nets across six datasets. The results indicate that the performance of each Des-Net is stable and generally improves with an increase in the number of parameters. 
This demonstrates the effectiveness and flexibility of the proposed LSAMD method for building variable-sized Des-Nets for various resource constraints.

\subsection{Ablation Studies}
\label{sec:ex_2}
As previously mentioned, there are two methods for propagating input batch data through each layer of the Ans-Net: 1) In the `Batch' method, the entire batch data is forwarded through either the base block or the dataset-related block, based on the probability value calculated by DAD. 2) In the `Img' method, each sample in the batch independently selects a block for propagation, either the base block or the dataset-related block.

We compared the performance of these methods across multiple datasets, as shown in Table \ref{tab:propagation_ways}. The `Batch' method generally outperforms the `Img' method on almost all datasets. Notably, the `Batch' method shows performance improvements of 19.78\% and 30.96\% over `Img' on the Food101 and Cars196 datasets, respectively. This improvement may be due to the `Img' method splitting the batch into two subsets, with only one subset used for the dataset-related blocks, resulting in less data for training these blocks compared to the `Batch' method. Consequently, `Img' tends to exhibit lower performance. The effect of the `Batch' and `Img' methods on base block selection is further discussed in the supplementary material.

\textbf{Various priority expansion stages}. 
% In the expansion phase, there are three potential priority positions: first, middle, and last. We conducted experiments to compare the impact of these positions on the performance of the Des-Nets.
% For a 6-layer Des-Net with learngene layers indexed at $(1, 4, 5, 7, 8)$, the indexes after applying first-stage, middle-stage, and last-stage priority expansions are $(1, \textbf{1}, 4, 5, 7, 8)$, $(1, 4, 5, \textbf{5}, 7, 8)$, and $(1, 4, 5, 7, 8, \textbf{8})$, respectively. The comparison results, as shown in Table \ref{lab_priority}, indicate that Des-Nets with last-stage priority expansion generally perform best across most datasets, while those with middle-stage priority expansion rank second.
% Based on the results of adding one layer, we further explore the priority of adding more layers. 
% Specifically, we show the results of 7-layer Des-Nets, and the indexes for priority expansions in middle-first-stage, last-first-stage, and last-middle-stage orders are $(1, \textbf{1}, 4, 5, \textbf{5}, 7, 8)$, $(1, \textbf{1}, 4, 5, 7, 8, \textbf{8})$, and $(1, 4, 5, \textbf{5}, 7, 8, \textbf{8})$, respectively. As shown in Table \ref{lab_priority}, the last-middle-stage priority expansion achieves the best results on all 7 datasets, while the last-first-stage priority expansion is second best on most datasets. These findings confirm that last-stage, middle-stage, and first-stage priorities lead to progressively better performance. 
In the expansion phase, there are three potential priority positions: first, middle, and last. Here, we conducted experiments to evaluate the impact of these positions on the performance of the Des-Nets.
For a 6-layer Des-Net with learngene layers indexed at $(1, 4, 5, 7, 8)$, the indexes after applying first-stage, middle-stage, and last-stage priority expansions are $(1, \textbf{1}, 4, 5, 7, 8)$, $(1, 4, 5, \textbf{5}, 7, 8)$, and $(1, 4, 5, 7, 8, \textbf{8})$, respectively. The results, presented in Table \ref{lab_priority}, show that Des-Nets with last-stage priority expansion generally outperform those with other priority positions across most datasets, while those with middle-stage priority expansion rank second.
Based on the results from adding one layer, we further investigate the priority of adding multiple layers. Specifically, we present the results for 7-layer Des-Nets, where the indexes for priority expansions in middle-first-stage, last-first-stage, and last-middle-stage orders are $(1, \textbf{1}, 4, 5, \textbf{5}, 7, 8)$, $(1, \textbf{1}, 4, 5, 7, 8, \textbf{8})$, and $(1, 4, 5, \textbf{5}, 7, 8, \textbf{8})$, respectively. As shown in Table \ref{lab_priority}, the last-middle-stage priority expansion yields the best results across all 7 datasets, while the last-first-stage expansion ranks second on most datasets. These findings confirm that expanding at the last stage, followed by the middle stage and then the first stage, leads to progressively improved performance.

\section{Conclusion}
This paper introduces LSAMD, a novel method for building variable-sized Des-Nets for different resource constraints. Initially, LSAMD integrates a dataset-specific block and a DAD module into each layer of the Ans-Net to construct the super Ans-Net. The super Ans-Net is then trained on multiple datasets, updating the parameters of the dataset-specific blocks while keeping the base block fixed. During training, the DAD module learns to search between the dataset-related and base blocks based on probabilities. The base blocks most frequently selected across datasets are extracted as learngene layers. These learngene layers are subsequently expanded to build Des-Nets of different scales, with priorities set in the order of last-middle-first. Experimental results on multiple datasets validate the superior performance of LSAMD.

\bibliography{main}

\appendix

\section*{Appendix}

We organize the appendix as follows.
\begin{itemize}
    \item In Section~\ref{hyper-p}, we provide the detailed hyter-parameters and hardware during training.
    \item In Section~\ref{datasets}, we list the information of the ten selected datasets for learngene extraction.
    \item In Section~\ref{select}, we show the impact of the `Batch’ and `Img’ methods on the base block selection in the supplementary material.
\end{itemize}

\begin{figure*}[t]
    \centering
    \includegraphics[width=\textwidth]{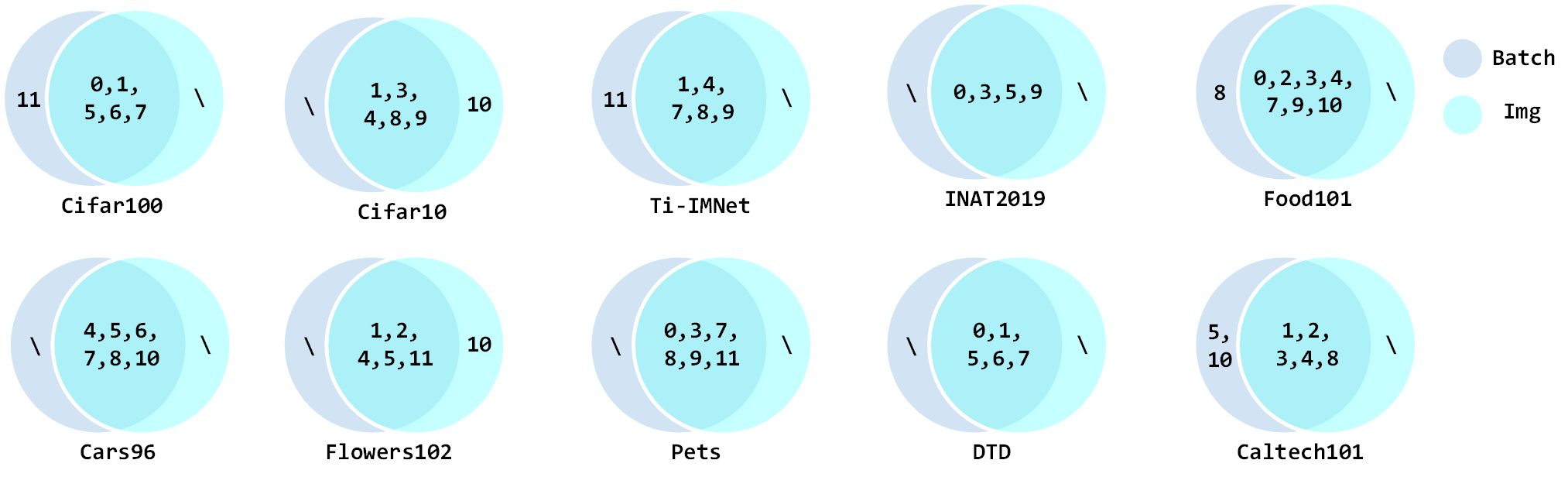}
    \caption[Figure]{The intersection of the indexes of the base blocks selected by the `Batch’ and `Img’ propagation methods. `\textbackslash’ means the null value.}
    \label{fig:intersecion}
\end{figure*}

\section{Hyper-parameters and Hardware}
\label{hyper-p}
We adopt a total batch size of 128 on 4 3090 GPUs. The initial learning rates for learngene extraction on each task and fine-tuning the built descendant models on the downstream datasets are shown in Table \ref{tab:learning_rate}.
In the above two training processes, the learning rate decay strategy is the poly method.

\begin{table}[htbp]
\begin{center}
\begin{tabular}{c|c|c}
\toprule[1pt]
Datasets         & Ans-Net  & Des-Net  \\ \midrule
Cifar100      & 5.00E-04 & 5.00E-04 \\
Cifar10       & 1.00E-06 & 5.00E-04 \\
Tiny-ImageNet & 1.00E-05 & 5.00E-04 \\
INAT2019      & 1.00E-05 & 5.00E-05 \\
Food101       & 1.00E-05 & 1.00E-05 \\
Cars196       & 5.00E-04 & 1.00E-05 \\
Flowers102    & 5.00E-04 & 1.00E-04 \\
Pets          & 1.00E-05 & 1.00E-04 \\
DTD           & 1.00E-04 & 1.00E-05 \\
Caltech101    & 1.00E-05 & 1.00E-05 \\
ADE-20K    & 1.00E-05 & 1.00E-05 \\
VOC12    & 1.00E-05 & 1.00E-05 \\
\bottomrule[1pt]
\end{tabular}
\caption{The initial learning rates for training the Ans-Net and fine-tuning the Des-Nets on ten datasets. `E-04’ represents $10^{-4}$.}
\label{tab:learning_rate}
\end{center}
\end{table}

\section{Datasets for Learngene Extraction}
\label{datasets}
We show the overview information of ten datasets for training the redesigned Ans-Net to extract learngene layers in Table \ref{tab:datasets}.

\begin{table}[t]
\begin{center}
\begin{tabular}{c|c|c|c}
\toprule[1pt]
Dataset       & Categories & Training & Testing \\ \midrule
CIFAR-100     & 100        & 50000            & 10000        \\
CIFAR-10      & 10         & 50000            & 10000        \\
Tiny-ImageNet & 200        & 100000           & 10000        \\
INAT2019      & 1010      & 268243           & 64401        \\
Food101       & 101        & 75750            & 25250        \\
Car196        & 196        & 8144             & 8041         \\
Flowers102    & 102        & 2040             & 818          \\
Pets          & 37         & 3680             & 3669         \\
DTD           & 47         & 3760             & 1880         \\
Caltech101    & 101        & 6942             & 1735            \\
ADE-20K &150  & 20210 & 200 \\
VOC12 &20 &10582 &1449 \\
\bottomrule[1pt]
\end{tabular}
\caption{The overview information of ten datasets for training the redesigned Ans-Net to extract learngene layers}
\label{tab:datasets}
\end{center}
\end{table}

\section{The Impact of the `Batch’ and `Img’ Methods on the Base Block Selection}
\label{select}
To verify the impact of the `Batch’ method on the base block selection, we also show the indices of base blocks selected in the `Batch’ and `Img’ propagation methods on all datasets in Figure \ref{fig:intersecion}. We can find that the base blocks selected by the both methods on each task are highly overlapped.
Although the `Img’ method has been proven to be valid by Mod-Squad, `Batch’ has better performance in training Ans-Net While achieving almost the same selection results as `Img’.
This further verifies the superiority of the `Batch’ in selecting base blocks.

\end{document}

\maketitle

\appendix

\section*{Appendix}

We organize the appendix as follows.
\begin{itemize}
    \item In Section~\ref{hyper-p}, we provide the detailed hyter-parameters and hardware during training.
    \item In Section~\ref{datasets}, we list the information of the ten selected datasets for learngene extraction.
    \item In Section~\ref{select}, we show the impact of the `Batch' and `Img' methods on the base block selection in the supplementary material.
\end{itemize}

\begin{figure*}[t]
    \centering
    \includegraphics[width=\textwidth]{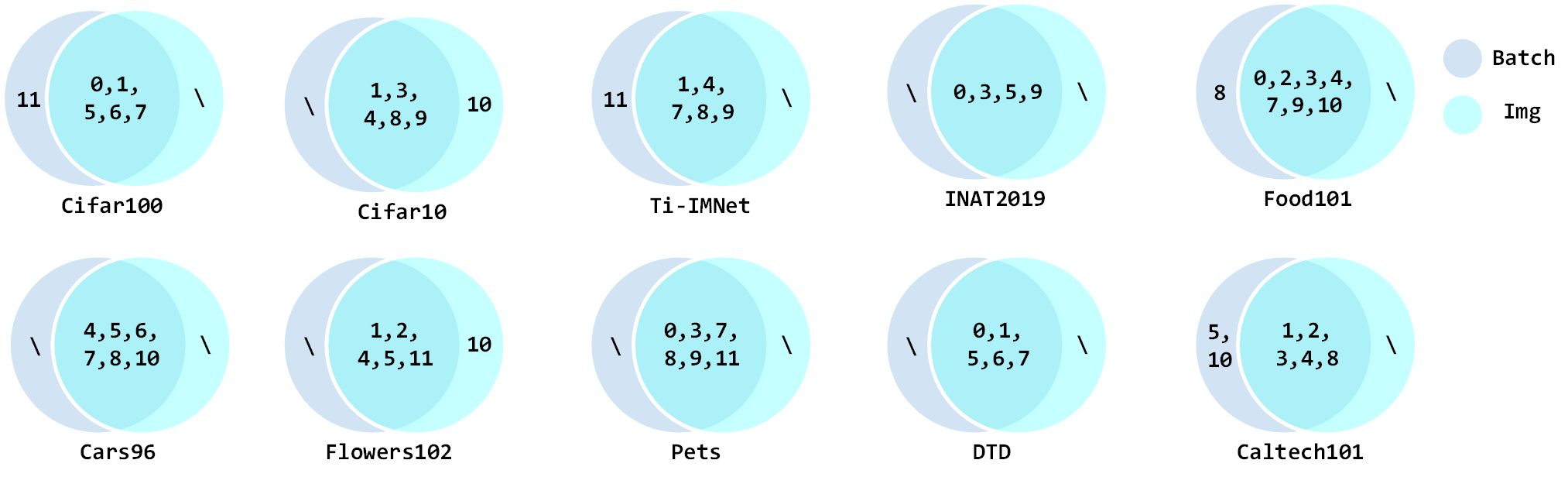}
    \caption[Figure]{The intersection of the indexes of the base blocks selected by the `Batch' and `Img' propagation methods. `\textbackslash' means the null value.}
    \label{fig:intersecion}
\end{figure*}

\section{Hyper-parameters and Hardware}
\label{hyper-p}
We adopt a total batch size of 128 on 4 3090 GPUs. The initial learning rates for learngene extraction on each task and fine-tuning the built descendant models on the downstream datasets are shown in Table \ref{tab:learning_rate}.
In the above two training processes, the learning rate decay strategy is the poly method.

\begin{table}[htbp]
\begin{center}
\begin{tabular}{c|c|c}
\toprule[1pt]
Datasets         & Ans-Net  & Des-Net  \\ \midrule
Cifar100      & 5.00E-04 & 5.00E-04 \\ 
Cifar10       & 1.00E-06 & 5.00E-04 \\ 
Tiny-ImageNet & 1.00E-05 & 5.00E-04 \\ 
INAT2019      & 1.00E-05 & 5.00E-05 \\ 
Food101       & 1.00E-05 & 1.00E-05 \\ 
Cars196       & 5.00E-04 & 1.00E-05 \\
Flowers102    & 5.00E-04 & 1.00E-04 \\ 
Pets          & 1.00E-05 & 1.00E-04 \\ 
DTD           & 1.00E-04 & 1.00E-05 \\
Caltech101    & 1.00E-05 & 1.00E-05 \\ 
ADE-20K    & 1.00E-05 & 1.00E-05 \\ 
VOC12    & 1.00E-05 & 1.00E-05 \\ 
\bottomrule[1pt]
\end{tabular}
\caption{The initial learning rates for training the Ans-Net and fine-tuning the Des-Nets on ten datasets. `E-04' represents $10^{-4}$.}
\label{tab:learning_rate}
\end{center}
\end{table}

\section{Datasets for Learngene Extraction}
\label{datasets}
We show the overview information of ten datasets for training the redesigned Ans-Net to extract learngene layers in Table \ref{tab:datasets}.

\begin{table}[t]
\begin{center}
\begin{tabular}{c|c|c|c}
\toprule[1pt]
Dataset       & Categories & Training & Testing \\ \midrule
CIFAR-100     & 100        & 50000            & 10000        \\ 
CIFAR-10      & 10         & 50000            & 10000        \\ 
Tiny-ImageNet & 200        & 100000           & 10000        \\ 
INAT2019      & 1010      & 268243           & 64401        \\ 
Food101       & 101        & 75750            & 25250        \\ 
Car196        & 196        & 8144             & 8041         \\
Flowers102    & 102        & 2040             & 818          \\
Pets          & 37         & 3680             & 3669         \\
DTD           & 47         & 3760             & 1880         \\ 
Caltech101    & 101        & 6942             & 1735            \\ 
ADE-20K &150  & 20210 & 200 \\
VOC12 &20 &10582 &1449 \\
\bottomrule[1pt]
\end{tabular}
\caption{The overview information of ten datasets for training the redesigned Ans-Net to extract learngene layers}
\label{tab:datasets}
\end{center}
\end{table}

\section{The Impact of the `Batch' and `Img' Methods on the Base Block Selection}
\label{select}
To verify the impact of the `Batch' method on the base block selection, we also show the indices of base blocks selected in the `Batch' and `Img' propagation methods on all datasets in Figure \ref{fig:intersecion}. We can find that the base blocks selected by the both methods on each task are highly overlapped. 
Although the `Img' method has been proven to be valid by Mod-Squad, `Batch' has better performance in training Ans-Net While achieving almost the same selection results as `Img'. 
This further verifies the superiority of the `Batch' in selecting base blocks.